\documentclass[letterpaper, 10 pt, conference]{ieeeconf}

\let\labelindent\relax
\usepackage{enumitem}
\setlist[enumerate]{itemsep=0mm}

\usepackage{url}            % simple URL typesetting
\usepackage{booktabs}       % professional-quality tables
\usepackage{amsfonts}       % blackboard math symbols
\usepackage{amssymb}
\usepackage{amsmath}
\usepackage{pdfpages}
\usepackage{comment}
\usepackage{mathtools}
\usepackage{placeins}
\usepackage{bbm}
\IEEEoverridecommandlockouts                              % This command is only needed if 
\usepackage{graphicx}
\usepackage{subcaption} %to have subfigures available
\usepackage{cases}
\usepackage{bm}
\usepackage{nicefrac}       % compact symbols for 1/2, etc.
\usepackage{microtype}      % microtypography
\usepackage{wrapfig}
\usepackage{xcolor}
\definecolor{darkgreen}{rgb}{0.0, 0.6, 0.13}

\usepackage{mathtools}
\usepackage{booktabs}       % professional-quality tables
\usepackage{bbm}
\usepackage{caption}

\DeclareCaptionFormat{withrule}{#1#2#3\hrulefill}
\DeclareCaptionFormat{withoutrule}{#1#2#3}
\usepackage{clrscode3e}
\DeclareMathOperator*{\argmin}{arg\,min}

\DeclareMathOperator*{\cost}{\mathcal{C}}
\DeclareUnicodeCharacter{2217}{*}

\usepackage{algorithmicx}
\usepackage[pagebackref=true,breaklinks=true,colorlinks,bookmarks=false]{hyperref}

\begin{document}

%%%%%%%%% TITLE
\title{Unsupervised Path Regression Networks}

\author{Michal Pándy$^{1}$, Daniel Lenton$^{2}$, Ronald Clark$^{2}$% <-this % stops a space
\thanks{$^{1}$Michal Pándy ({\tt\small mp988@cam.ac.uk}) is at the Department of Computer Science and Technology,
        University of Cambridge.} %
\thanks{$^{2}$Daniel Lenton ({\tt\small daniel.lenton11@imperial.ac.uk}) and Ronald Clark ({\tt\small ronald.clark@imperial.ac.uk}) are at the Department of Computing, Imperial College London.}%
}

\maketitle
%\thispagestyle{empty}

% Motion planning is a fundamental problem in robotics and machine perception. Sampling-based planners find accurate solutions by exhaustively exploring the space, but are inefficient and tend to produce jerky motions. Optimization and learning-based planners are more efficient and produce smooth trajectories. A significant hurdle that learning-based path planning approaches face lies in constructing a differentiable cost function that simultaneously minimizes path length and avoids collisions. These two objectives are conflicting by nature -- path length is continuous and well-behaved, but collisions are discrete non-differentiable events. Reconciling these terms has been a significant challenge in optimization-based motion planning. The main contribution of this paper is a novel cost function that guarantees collision-free shortest paths are found at its minimum. We show that our approach works seamlessly with RGBD input and predicts high-quality paths in 2D, 3D, and 6 DoF robotic manipulator settings. Our method also reduces training and inference time compared to existing approaches, in some cases by orders of magnitude.

%Motion planning problems are conventionally solved in an iterative manner, by either discretizing the space and performing a graph search, or by sampling the continuous state space and attempting to connect nodes. Recent iterative learning methods have greatly improved inference time, but require ground truth paths for training, which are very costly to compute. 

\begin{abstract}
We demonstrate that challenging shortest path problems can be solved via direct spline regression from a neural network, trained in an unsupervised manner (i.e. without requiring ground truth optimal paths for training). To achieve this, we derive a geometry-dependent optimal cost function whose minima guarantees collision-free solutions. Our method beats state-of-the-art supervised learning baselines for shortest path planning, with a much more scalable training pipeline, and a significant speedup in inference time.
\end{abstract}

% For example, consider a robotic arm that has to pick up a cup on a cluttered table; the arm has to rapidly plan a path from its start position to the cup without bumping into any other objects in the vicinity. Another example is a quadcopter surveying a warehouse; the quadcopter needs to rapidly plan to avoid moving people and objects while covering all areas of the warehouse before the battery running out. 

%\saythanks
\section{Introduction}
\label{intro}
Motion planning is essential for most robotics and embodied AI applications, but is also an exceptionally difficult problem for multiple reasons. Firstly, the planner is often required to find paths of minimal length in order to minimize power consumption and execution time. Secondly, a usable path must avoid obstacles (taking quadcopter as an example, a collision might cause fatal damage). Thirdly, the motion of many real systems (e.g., robot arms) is limited by the controllable actuators and this limits the set of feasible trajectories. Finally, real-world robots are limited to partial observations of their surroundings, acquired from on-board sensors.

\begin{figure}[t!]
    \begin{center}
    \includegraphics[width=\columnwidth]{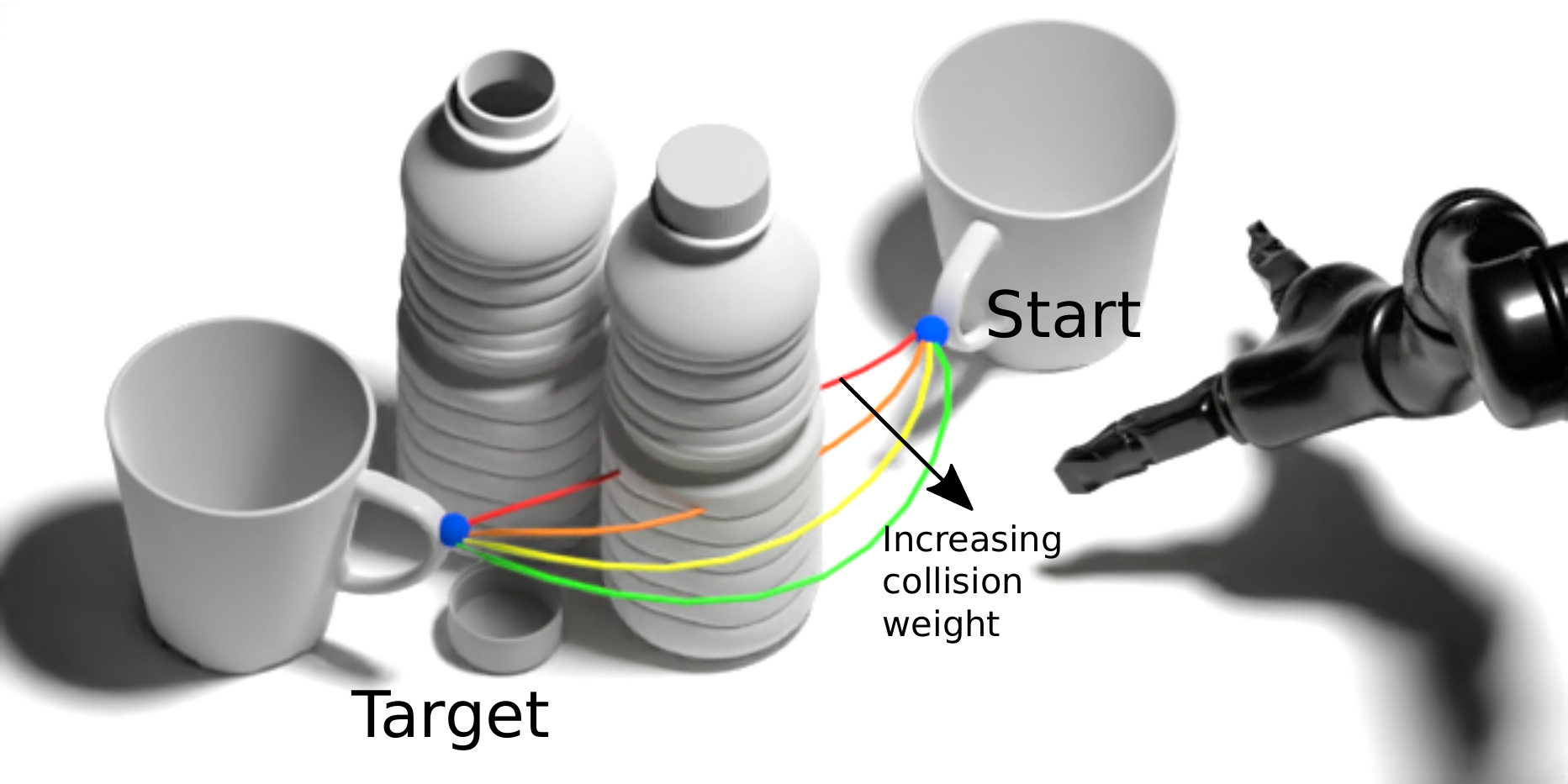}
    \caption{\textbf{Path length and collision cost are conflicting objectives.} Here, we show an application of motion planning where a robot arm needs to estimate a path to move the cup from a start configuration to a target configuration. Most planning methods tune a weighted combination of path length $l$ and collision $c$ to find a path. In this paper, we design $l$ and $c$ to guarantee collision-free paths to be found at the optimum of the cost function and avoid the need for a weighting between the two objectives.}
    \label{fig:cover}
    \end{center}
\end{figure}

Existing methods based on sampling, grid, or tree searches successfully avoid obstacles by querying points in the configuration space and checking whether collisions occur. These approaches are accurate and have high success rates, but their run-time can be prohibitive. These limitations are addressed by \textit{gradient-based planners}, which can efficiently find smooth trajectories. However, formulating a suitable cost function for these approaches is challenging as the two terms, collision cost and path length, are inherently conflicting. The collision cost is a hard binary constraint that is non-differentiable by nature. The most common approach to address this is to relax the collision cost by a soft signed distance function, but this has two major disadvantages: 
\begin{enumerate}
\item The relaxed cost function does not guarantee the shortest paths to be found at its optimum. 
\item A hyperparameter which trades between collision and path length needs to be tuned (see Fig.~\ref{fig:cover}). 
\end{enumerate}

In this paper we propose to train a network that \textit{directly regresses} an entire path from start to goal in a single forward pass, by minimising an unsupervised \textit{novel cost function} at training time. The cost function we propose is similar in form to those in the optimization-based planning literature \cite{zucker2013chomp, mukadam2018continuous}, but unlike existing methods, our novel formulation guarantees \textit{collision-free shortest paths} to be found at the minima. Our cost function does not contain any hyperparameter, simplifying the learning process. The trained network is conditioned on some form of scene description, but our method does not limit the parameterization of the scene description. For example, we show our method works with scenes specified as a list of objects (their locations and shapes), and also with scenes parameterized as an RGBD image. This makes our approach applicable for sim-to-real learning, where we can leverage the full state of the simulator to construct the optimal loss at training time, to train a network which only receives image observations at inference time.

We demonstrate state-of-the-art performance for learning-only approaches on benchmark tasks, including point-mass path-planning in 3D space and reaching target joint configurations in the presence of obstacles for a 6-DoF robotic manipulator. Importantly, our method does not need to pre-compute a dataset of shortest paths for training, so we are also able to reduce \textit{total} training time by almost two orders of magnitude compared to supervised approaches.
\section{Related Work}

For low-dimensional problems ($< 3$ DoF), \textbf{graph-based planners} are efficient and can find optimal solutions. These approaches construct a graph by discretizing the space and connecting neighboring cells. The shortest path can then be found using variations of dynamic programming \cite{dijkstra1959note, hart1968formal, hart1972correction, dechter1985generalized}. However, these approaches quickly become intractable when moving to higher dimensions. 

For a problem with more degrees of freedom ($> 3$ DoF), \textbf{sample-based planners} such as rapidly-exploring random tree (RRT) and probabilistic roadmap (PRM) are the most popular approaches. These planners dynamically build a network of paths at run-time by attempting to connect nodes which are sampled in continuous space \cite{amato1996randomized, kuffner2000rrt, karaman2011sampling, klemm2015rrt}. Although these methods exhibit probabilistic convergence guarantees, their runtime performance is prohibitive for many real-world applications, and the paths produced are generally jerky, requiring post-processing.

Continuous planners that do not rely on discretizing the space can find smooth solutions more efficiently. \textbf{Potential field planners}, for example, model obstacles as a repulsive force and model path length as an attractive force \cite{quinlan1993elastic, brock2002elastic,garber2004constraint}. However, these methods easily get stuck in local minima and typically require a good initial path estimate from sample-based planners. 

Advances in \textbf{optimization}-based planners \cite{zucker2013chomp, kalakrishnan2011stomp, park2012itomp,schulman2014motion} have demonstrated that paths can be optimized directly from naive initial guesses, with no sample-based planners involved. Apart from the difficulty with tuning a cost function for a specific scene, the requirement for multiple gradient steps at inference time can be prohibitive in dynamic contexts requiring fast robotic responses.

\textbf{Deep learning}-based methods such as \cite{qureshi2019motion, srinivas2018universal, bency2019neural,bhardwaj2020differentiable,tamar2016value} train networks to iteratively predict trajectories that bring the agents closer to target state. These methods use ground-truth paths computed using a standard planner to serve as training examples. Apart from the significant computational overhead associated with generating such ground-truth paths, these approaches may also be susceptible to biases created by employing traditional planners to generate the training samples \cite{jurgenson2019harnessing}.

The need for ground-truth paths can be overcome by using \textbf{reinforcement learning}. \cite{faust2018prm,mnih2016asynchronous, jaderberg2016reinforcement, levine2016end, aljalbout2020learning}. However, reinforcement learning-based approaches are often very sample-inefficient, particularly when learning from sparse rewards, requiring many trials to train. These approaches can also often struggle to generalize between different tasks and environments.

\section{Approach}

\begin{figure}[h!]
    \centering
    \includegraphics[width=\columnwidth]{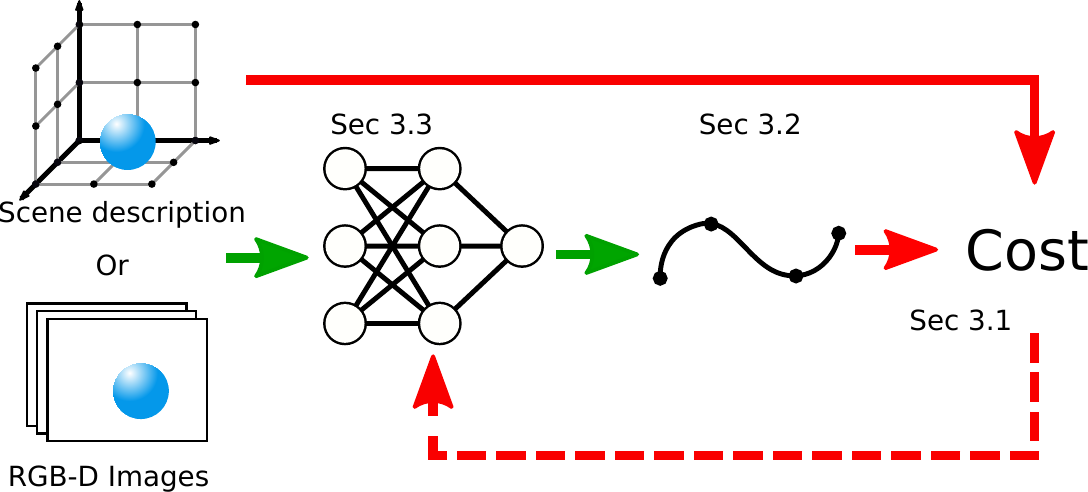}
    \caption{\textbf{Overview of our method.} The \textbf{\color{darkgreen} solid green} arrows indicate data flow at inference time. The \textbf{\color{red} solid red} arrows indicate data flow at training time. The \textbf{\color{red} dotted red} arrows indicate gradient flow, which is used to update the weights of the network. If a scene description is available at run-time, the weights can also be updated at test time to refine the path. }
    \label{fig:app-overview}
\end{figure}

In this section, we describe our approach for learning to find shortest collision-free paths. A high-level overview of our approach is illustrated in Figure \ref{fig:app-overview}. We use a neural network (described in Section \ref{sec:network}) to regress a path parameterized as a spline (further described in Section \ref{sec:parametrisation}). Finally, we calculate the path's cost using our novel cost function (derived in Section \ref{sec:cost}) and update the network weights using stochastic gradient descent. In the next sections, we give a detailed description of each of these three components.

%The visualisation of the overview mentioned above can be seen in Figure \ref{fig:app-overview}.

\subsection{Cost function derivation} 
\label{sec:cost}

In this section, we outline the full derivation of our cost function. As mentioned before, the critical challenge in formulating a smooth cost function for shortest path planning lies in the fact that collision avoidance is a hard constraint, which is often replaced by soft penalty terms. Finding a weighting between collision and path length terms is challenging, as they are not directly comparable. For this reason, standard optimization-based methods \cite{schulman2014motion, zucker2013chomp} require per-task or per-scene calibration, limiting their ability to generalize across diverse sets of scenes. Hence, we derive a novel formulation that guarantees collision-free paths, invariant to scene scaling. 

Let $O$ be a set of arbitrary obstacles, $\Pi$ be a set of corresponding obstacle observations, $s \in \mathbb{R}^{d}$ be a start configuration, $t \in \mathbb{R}^{d}$ be a target configuration, and $\theta_{p} \in \Theta_{p}$ a path parameterization. We aim to optimize the path planning function:
\begin{equation}
f_{\theta}(\Pi, \: s, \: t) = \theta_{p} 
\end{equation}

To successfully optimize $f_{\theta}$, we define a loss function:
\begin{equation}
\cost(O, \: \theta_{p})
\end{equation}

In our work, we aim to optimize $f_{\theta}$ to converge to paths that are shortest and collision-free. This assumption naturally leads to $\cost$ having components penalising collisions and path lengths. Hence, we write: 
\begin{equation}
\cost(O, \: \theta_{p}) = l(\theta_{p}) \: + \: c(O, \: \theta_{p})
\end{equation}

where $l$ and $c$ are \textit{length} and \textit{collision penalty} measures, respectively. While the exact structure of $c$ is unknown for now, we can already point out useful properties it should have. Suppose our optimisation problem has an optimal path parameterized by $\theta_{opt} \in \Theta_{p}$. Then, we require:

\begin{itemize}
	\item \textbf{Minimum property (MP)}: 
	\begin{equation*} \forall \: \theta_{p} \in \Theta_{p}: \: c(O, \: \theta_{opt}) \leq c(O, \: \theta_{p}) \end{equation*}

    \item \textbf{Non-colliding property (NP) \label{nc-prop}}: 
     \begin{equation*} 
       \forall \: \theta_{p} \in \Theta_{p}: \theta_{p} \: \text{does not collide}  \Leftrightarrow  c(O, \: \theta_{p}) = 0 
     \end{equation*}
\end{itemize}
Further, since we require that $\cost$ is a valid cost function, we have:

\begin{itemize}
     \item \textbf{Global optima property (GP)}: 
     \label{glob-optim}
     \begin{align*}
         \forall \: \theta_{p} \in \Theta_{p}: \cost(O, \: \theta_{opt}) & \leq \cost(O, \: \theta_{p}) \\
        \llap{$\Leftrightarrow$ \qquad}  l(\theta_{opt}) \: + \: c(O, \: \theta_{opt})  & \leq l(\theta_{p}) \: + \: c(O, \: \theta_{p}) \\
        \llap{$\Leftrightarrow$} \qquad l(\theta_{opt}) \: - \: l(\theta_{p}) & \leq \: c(O, \: \theta_{p})
     \end{align*}
\end{itemize}

In general, we have an underlying assumption that a non-colliding solution to the planning problem exists. We propose that any formulation of $\cost$ with $c$ satisfying the three properties above, will have its minima in paths which are non-colliding and shortest possible. To show that $\theta_{opt}$ is non-colliding, assume for contradiction that $\theta_{opt}$ is colliding and an arbitrary $\theta' \in \Theta_{p}$ is non-colliding. Then, by NP we have $c(O, \: \theta_{opt}) > 0$, and so by MP necessarily also $c(O, \: \theta') > 0$. This by NP however means that $\theta'$ is colliding and we have a contradiction. To show that $\theta_{opt}$ is shortest possible, we show that $\forall \theta' \in \Theta_{p}: \: l(\theta') < l(\theta_{opt}) \Rightarrow \theta' \: \text{collides}$. Hence, for arbitrary $\theta' \in \Theta_{p}$ assume that $l(\theta') < l(\theta_{opt})$. Then by GP and our assumption, we have $0 < l(\theta_{opt}) - l(\theta') \leq c(O, \: \theta')$. By NP and $c(O, \: \theta') > 0$ we finally have that $\theta'$ collides.

Now, to define a loss function $\cost$ which we can optimize, we have to define $c$ that satisfies NP, MP, GP. One option for picking $c$ is:

\begin{equation}
    c(O, \theta_{p}) = \sum_{o \in O} 1_{o}(\theta_{p}) * R(o)
\end{equation}
\begin{equation}
    1_{o}(\theta_{p}) =
    \begin{dcases*}
    \: 1
       & if  path given by $\theta_{p}$ collides with $o$ \\[1ex]
    \: 0
       & otherwise
    \end{dcases*}
\end{equation}

where $R(o)$ is defined to be the bounding sphere circumference of object $o \in O$. It is possible to formally show that this formulation satisfies NP, MP, and GP. The intuition behind the proof is that object circumference collision penalties always yield non-colliding minimal paths, as the path can simply travel around the obstacle to minimise the cost. Therefore, the loss function that we propose takes the form:
\begin{equation}
\cost(O, \: \theta_{p}) = l(\theta_{p}) \: + \: \sum_{o \in O} 1_{o}(\theta_{p}) * R(o)
\end{equation}

\subsection{Path Parameterization} 
\label{sec:parametrisation}
In the previous section, we describe a cost function $\cost$ which we can use to optimize a planning function $f_{\theta}$ not requiring any additional hyperparameter tuning. In this section, we consider a possible parameterization of $\theta_{p}$ so that $\cost$ is differentiable, and we may optimize $f_{\theta}$ using stochastic gradient descent.

As noted in the introduction, we are aiming to train path regression networks. In contrast to iterative approaches such as \cite{qureshi2019motion, tamar2016value}, which require input from the previous agent state $s_{t} \in \mathbb{R}^{d}$ to infer the next state $s_{t + 1} \in \mathbb{R}^{d}$, we instead use $f_{\theta}$ to predict paths from the start configuration to the goal configuration in one inference step. This way, we can ensure a good $f_{\theta}$ inference speed both at train and at test time. While letting $\theta_{p} \in \Theta_{p}$ be a fixed sized unrolling of states in the task space is possible, this would require a large cardinality of $\theta_{p}$, for $f_{\theta}$ to be able to express complex smooth paths. This approach in practice is hard to optimize and would incur significant inference speed penalties. As with other works \cite{judd2001spline, magid2006spline, yang2014spline, mercy2017spline}, we consider a parameterization where $\theta_{p} \: = \: \{ (p_{k}, w_{k}) \: | \:\: p_{k} \in \mathbb{R}^{d}, w_{k} \in [0, 1], k \in \{1, 2, ..., n\} \}$ and $n \in \mathbb{N}_{>0}$ is a problem specific task complexity parameter. With such parameterization, we can use $\theta_{p}$ to define a path in the form of a non-uniform rational B-spline (NURBS) with control points $p_{k}$, control point weights $w_{k}$, a default open-uniform knot vector to anchor the spline in the start and goal configurations, and a degree parameter $p \in \mathbb{N}_{>0}$. In practice, $p > 1$ is sufficient for most setups.

Now, for an arbitrary $\theta_{p} \in \Theta_{p}$ and object set $O$ we show how to approximate $\cost(O, \: \theta_{p})$ using our NURBS parameterization. We achieve this by evaluating the $\theta_{p}$ NURBS interpolation with a high enough sampling rate $1 / s$ for each value in $B \coloneqq \{ s * k \: | \: 0 \leq s * k \leq n - p \:, \: k \in \mathbb{N} \}$. Let $N : \Theta_{p} \times B \longrightarrow \mathbb{R}^{d}$ be the NURBS interpolation.

In case of the length component $l(\theta_{p})$, we have:
\begin{equation}
\begin{aligned}
l(\theta_{p}) &= \int_{0}^{n-p}\left \|  N(\theta_{p}, x) \right \| dx 
= \lim_{\delta x \rightarrow 0}\sum_{x=0}^{n-p}\left \| N(\theta_{p}, x) \right \|\delta x \\ 
&\approx \sum_{x \in \{0, s, 2s, ...\}}^{n-p-s}\left \| N(\theta_{p}, x + s) - N(\theta_{p}, x)\right \|\
\end{aligned}
\end{equation}

In case of the collision component $c(O, \: \theta_{p})$, we first define an object selector function: 

\begin{equation}
\label{eq:selector}
    \tau(O, X_{p}) = \argmin_{o \in O} \: SDF(o, X_{p})
\end{equation}

Where $X_{p} \in \mathbb{R}^{d}$ and $SDF$ is a differentiable signed distance function. Now, we define a point cost function:
\begin{equation}
\label{eq:point-cost}
    \hat{c}(X_p, O, \theta_p) = 
    \begin{cases}
     \frac{R(\tau(O, X_p))}{\Delta(X_p, O, \theta_p)}       & SDF(\tau(O, X_p), X_p) < 0 \\
                                                         0  & otherwise 
    \end{cases}
\end{equation}

with $\Delta$ providing the number of configurations along $\theta_{p}$ which collide with the same object as a given configuration, simply defined as:
\begin{equation}
    \Delta(X_{p}, O, \theta_{p}) = \sum_{X \in \{0, s, 2s, ...\}}^{n-p} \delta_{\tau(O, X_{p})}^{\tau(O, N(\theta_{p}, X))}
\end{equation}

\noindent Note that $\Delta$ is always greater than $0$, due to the branching condition in $c_{p}$, as every colliding point has at least itself as a corresponding colliding point with the same object. Hence, we can finally write $c$ under the NURBS parameterization as:

\begin{equation}
\label{cost-nurbs}
\begin{aligned}
c(O, \theta_{p}) &=
\sum_{o \in O} 1_{o}(\theta_{p}) * R(o)\\ &= {\sum_{x \in \{0, s, 2s, ...\}}^{n-p} \hat{c}(N(\theta_{p}, x), O, \theta_{p})}
\end{aligned}
\end{equation}

Although we can now easily compute $c$ using NURBS parameterized $\theta_{p}$, we can not use gradient descent to optimize $f_{\theta}$ using $c$ just yet, as the gradients of $c$ are undefined. To provide gradients for $c$, we further upper bound it as:
\begin{equation*}
\label{eq:nurbs-coll}
\sum_{x \in \{0, s, 2s, ...\}}^{n-p} \hat{c}(N(\theta_{p}, x), O, \theta_{p}) * H(\min_{o \in O} \: SDF(o, N(\theta_{p}, x)))
\end{equation*}
\begin{equation}
H(x) = \frac{2}{1 + e^{x - \delta}}
\end{equation}

\noindent where $H: \mathbb{R} \longrightarrow \mathbb{R}$ is a smooth approximation of a step function and $\delta$ is a safe distance parameter, which controls the extent to which the paths should avoid the obstacles. $H$ could in practice be any function with $H(\delta) = 1$, $\forall x \leq \delta: \: H(x) \geq 1$, and $\lim_{x\rightarrow \infty} H(x) = 0$. The intuition behind the given approximation of $c$ lies in the fact that $\hat{c}$ provides the scaling of the gradient that ensures obstacle avoidance, while the gradient of $H$ directs path points outside of objects. Note that $\delta$ is not a parameter intended to be tuned, but rather a way to control how far optimal solutions should lie from objects. Further, note that although we derived an approximation to the optimal cost function from Section \ref{sec:cost}, the approximate collision cost is at least the true collision cost and the approximate length cost is at most the true length cost. This ensures that in the approximate setting, optimal paths are guaranteed to be non-colliding.

\subsection{Network}
\label{sec:network}
\begin{figure}[!htpb]
    \centering
    \includegraphics[width=90
    mm]{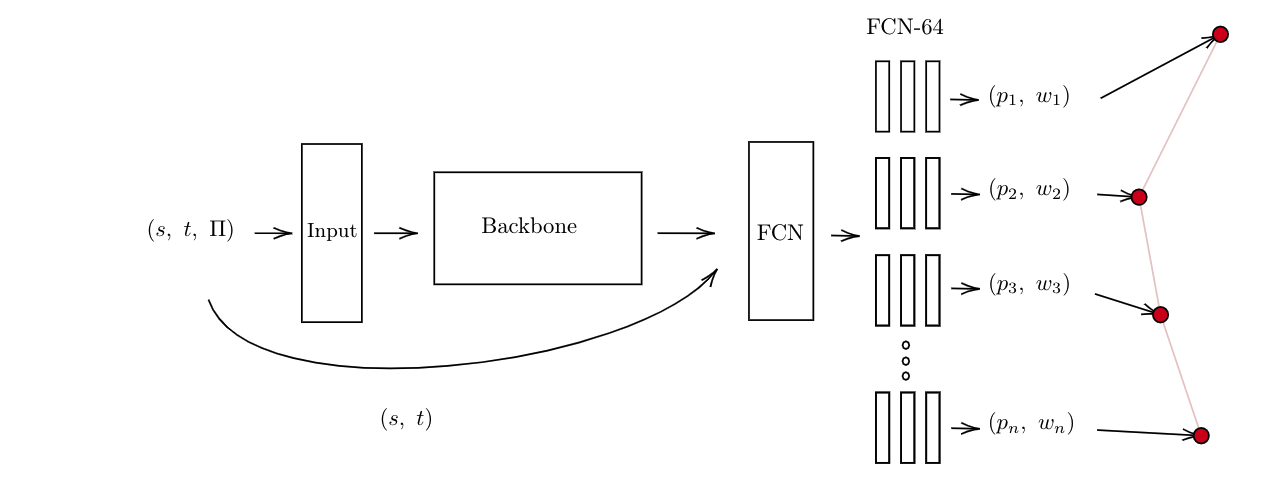}
    \caption{General network architecture used in our experiments.}
    \label{fig:plan-arch}
\end{figure}

The network architecture we use in our approach depends on the particular planning domain. In case of planning from images (\ref{sec:partial-obs}), we use a convolutional input layer to process the RGBD images, followed by a ResNet50\cite{he2016deep} backbone. In case of 6 DoF (\ref{sec:dof}) and 3D planning (\ref{sec:cont-plan}), we utilise vectorized scene descriptions (these descriptions are $\in \mathbb{R}^{k \times d}$, where $k$ is the obstacle count and $d$ is the dimension of the obstacle properties) which are processed by a fully connected input layer, followed by $10$ highway layers \cite{srivastava2015highway}. The output layer in general consists of $n$ fully connected networks for each spline anchor point. Our architecture is visualised in \autoref{fig:plan-arch}.

% TOOD: Insert network figure

\section{Datasets and baselines}

We compare our approach against representative sampling-based planners, an optimization-based planner, and a learning-based planner.

\noindent \textbf{RRT*} \cite{karaman2011sampling}, \textbf{Informed-RRT*} \cite{gammell2014informed}, and \textbf{BIT*} \cite{gammell2015batch}: are perhaps the most widely used sampling-based planning algorithms in use today. These methods are optimized versions of RRT that guarantee to find the shortest path when run for an indefinite amount of time. \\
\noindent \textbf{CHOMP} \cite{zucker2013chomp}: is a well-performing gradient-based motion planning algorithm. Similar to ours, CHOMP's cost function has terms resembling our length term and collision term, scaled by a hyperparameter. 

\noindent \textbf{MPNet} \cite{qureshi2019motion}: is a state-of-the-art learning-based planner. Given a point cloud scene representation with the current agent state, MPNet outputs the next agent state that will bring it closer to the goal configuration. The MPNet method further employs lazy state contraction and re-planning, which are algorithmic methods for refining the paths. In our experiments however, we focus the learning-based components of MPNet, as our method can be easily extended with algorithmic path corrections such as re-planning, and these algorithmic corrections are generally not applicable in partially observable environments. 

We test our approach using both synthetic and real-world data. Specifically, we use the following six datasets.

\noindent \textbf{simple-2D}: We randomly sample a rectangle and a sphere in a 2D scene, together with a start and target position, such that a straight line path would collide with either of the objects. This simple dataset is only used for comparing the characteristics of our cost function to others and to give an intuitive visualisation.

\noindent \textbf{Complex3D} \cite{qureshi2019motion}: This dataset contains 110 scenes with 5000 near-optimal paths generated using RRT* (note, unlike \cite{qureshi2019motion} our approach does not need these paths for training). The training split contains 100 scenes with 4000 ground truth paths. The testing split consists of 100 scenes (contained in the training set) but with 200 unseen paths. There is also a test set of 10 unseen environments with 2000 paths.

\noindent \textbf{Table-top shapes}: We generate a table-top RGBD dataset using CoppeliaSim\cite{coppeliaSim} by randomly placing floating cuboids, cylinders, and spheres such that they intersect with the ground plane of a large bounding cuboid. They are also permitted to intersect with each other. We randomized camera positions and focal lengths for each image, with a bias to face towards the ground plane, where the objects are spawned. We plan to release this dataset for reproducibility and to allow others to train and benchmark their approaches.

\noindent \textbf{RGB-D Scenes Dataset v.2}: \cite{lai2011large}: This dataset contains RGBD images of real-world table-top scenes that we use for testing our approach.

\noindent \textbf{all-6DoF} and \textbf{difficult-6DoF}: We generate these datasets for comparing our method on 6 DoF robotic manipulator planning problems. The datasets assume a 6 DoF Kinova Mico\cite{campeau2019kinova} manipulator tasked to reach specified target configurations in the presence of a fixed-sized box obstacle of dimensions $0.2\text{m}\text{ x }0.2\text{m x }0.2\text{m}$, $0.29\text{m}$ away from the robot base. For \textbf{all-6DoF}, we sample random start and target manipulator configurations such that these configurations do not collide with the box. For \textbf{difficult-6DoF}, we likewise sample such configurations, but with the additional constrain that a linear interpolation in the start \& target join angles does not solve the planning problem.

\section{Evaluation}

In this section, we evaluate our cost function together with the proposed parameterization in various domains. Our goal is to focus on answering the following:

\begin{enumerate}
    \item How does our cost function perform in comparison to related methods? (\ref{eval:chomp})
    \item Does our method perform up to par with state-of-the-art approaches when planning from full scene descriptions and from images? (\ref{sec:cont-plan}, \ref{sec:partial-obs}) 
    \item How does our method perform in higher-dimensions with robotic manipulators? (\ref{sec:dof})
\end{enumerate}

\subsection{Cost function evaluation}
\label{eval:chomp}
In this section, we assess how our cost, $\cost$, compares to the CHOMP collision cost \cite{zucker2013chomp}. For a single sample point $x \in \mathbb{R}^{d}$, the CHOMP collision term is as follows, with $\varepsilon \in \mathbb{R}$ being a calibrated constant:
\begin{equation}
c_{CHOMP}(x) =
\begin{dcases*}
\: -SDF(x) + \frac{1}{2} \varepsilon
   & if $SDF(x) < 0$
   \\[1ex]
\: \frac{1}{2\varepsilon} (SDF(x) - \varepsilon)^2
   & if $0 < SDF(x) \leqslant \varepsilon$
   \\[1ex]
\: 0
   & otherwise
\end{dcases*}
\end{equation}
We choose to compare the cost functions on \textit{simple-2D} in order to make brute-force optimization tractable.

\begin{figure*}[h!]
    \centering
    
    \begin{subfigure}[t]{0.49\textwidth}
         \raisebox{-\height}{\includegraphics[width=0.49\textwidth,height=0.49\textwidth]{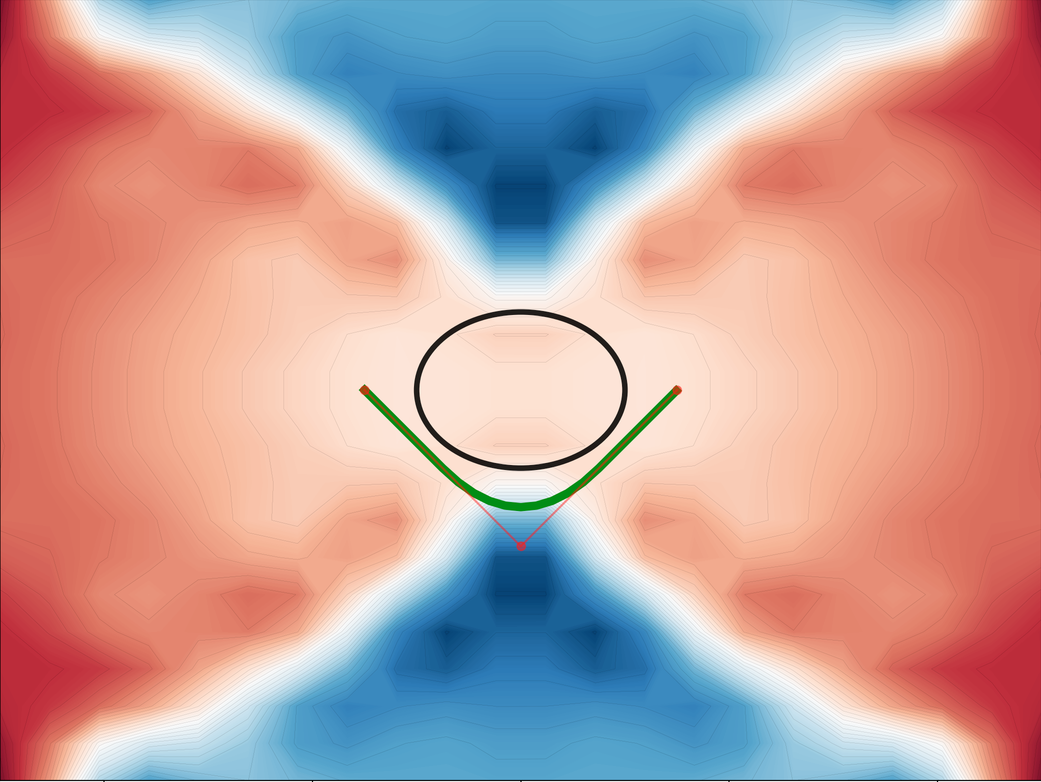}}
         \raisebox{-\height}{\includegraphics[width=0.49\textwidth,height=0.49\textwidth]{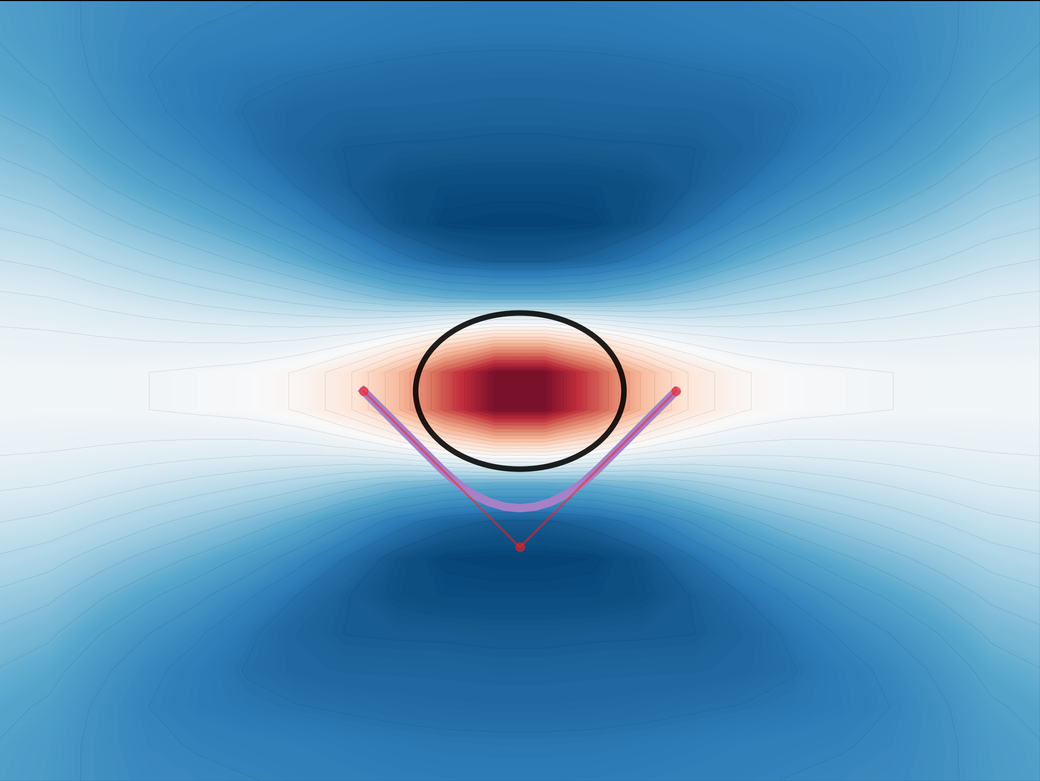}}%
         \caption{Sphere planning problem that was used to calibrate the CHOMP cost.}
         \label{fig:one-a}
    \end{subfigure}
    \hfill
    \begin{subfigure}[t]{0.49\textwidth}
         \raisebox{-\height}{\includegraphics[width=0.49\textwidth,height=0.49\textwidth]{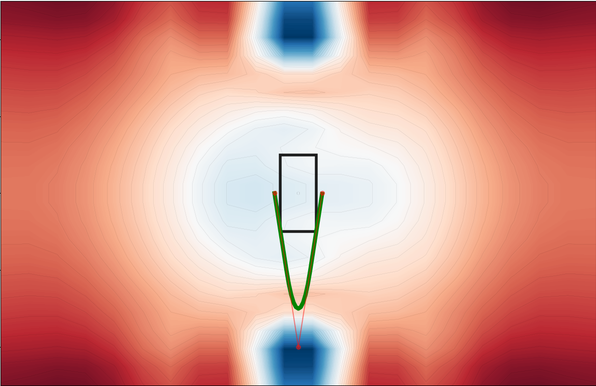}}
         \raisebox{-\height}{\includegraphics[width=0.49\textwidth,height=0.49\textwidth]{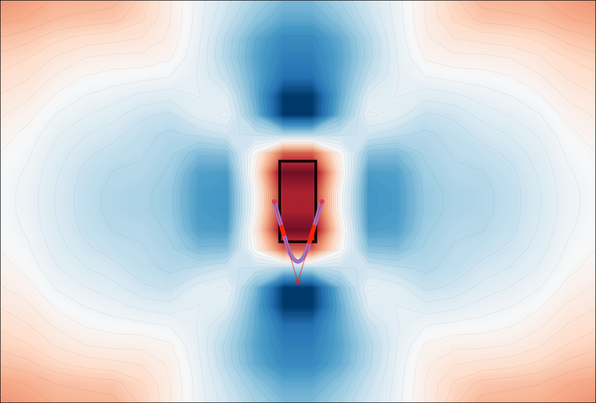}}%
         \caption{Failure case which yields collisions for the calibrated CHOMP cost.}
    \end{subfigure}
    %%%%%%%%%%%%%%%%%%%%%%%%%%%%%%%%%%%%second row
    \begin{subfigure}[t]{0.49\textwidth}
        \raisebox{-\height}{\includegraphics[width=0.49\textwidth,height=0.49\textwidth]{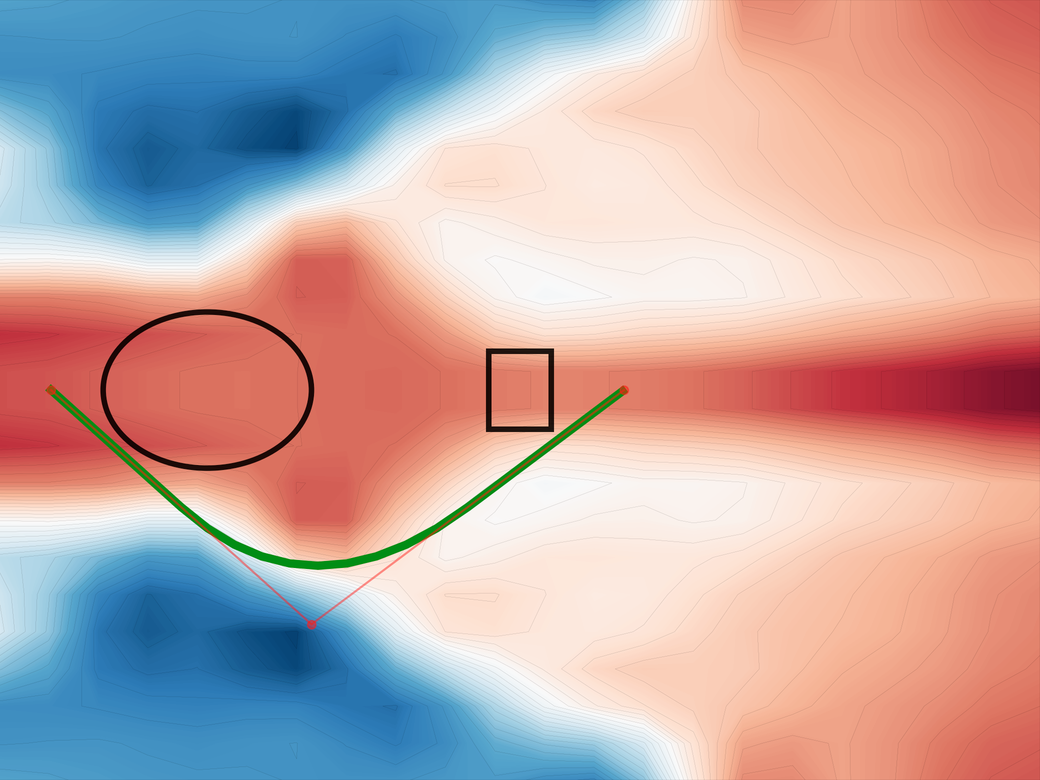}}
        \raisebox{-\height}{\includegraphics[width=0.49\textwidth,height=0.49\textwidth]{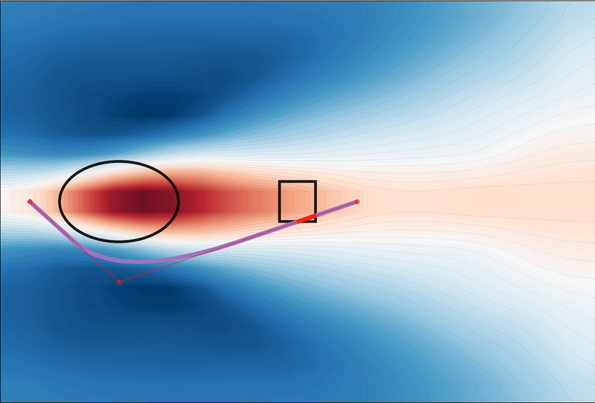}}%
        \caption{$\cost$ successfully avoids rectangle corners.}
    \end{subfigure}
    \hfill
    \begin{subfigure}[t]{0.49\textwidth}
          \raisebox{-\height}{\includegraphics[width=0.49\textwidth,height=0.49\textwidth]{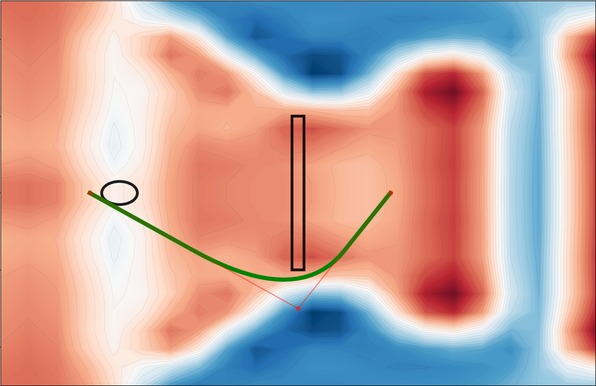}}
          \raisebox{-\height}{\includegraphics[width=0.49\textwidth,height=0.49\textwidth]{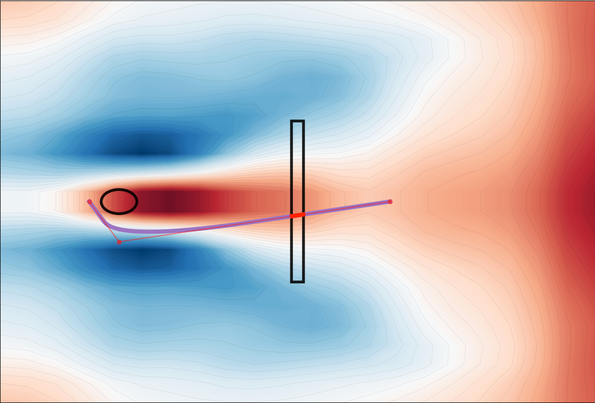}}%
          \caption{$\cost$ successfully avoids thin rectangles.}
    \end{subfigure}
    
    \caption{
    \textbf{Comparing our cost function to that of CHOMP}\cite{zucker2013chomp}. In each image pair, we show the result of optimizing $\cost$ (green) on the left, and the result of optimizing the CHOMP collision cost (purple) on the right. In all examples, we use a single control point NURBS parameterization. The background heat maps represent the cost function's values at different control point positions, with red regions being the maxima and blue areas being the minima. The paths obtained by minimizing $\cost$ are collision-free with the shortest possible lengths for the given number of control points.}
    \hrulefill
    \label{fig:one}
\end{figure*}

\noindent \textbf{Setup:} As a first step, we calibrate $\lambda$ (collision weight hyperparameter) and $\varepsilon$ in the CHOMP collision cost for a simple sphere problem in our dataset, as seen in Figure \ref{fig:one-a}. We perform this calibration so that the optimal CHOMP cost path is collision-free, with the same length as our cost's optimal path. 

\noindent \textbf{Results:} In this setup, out of 150 planning problems, $\cost$ achieves a 100\% success rate, while the calibrated CHOMP\cite{zucker2013chomp} cost achieves a 79.33\% success rate, and an uncalibrated CHOMP\cite{zucker2013chomp} cost, with default $\lambda = 1$, $\varepsilon = 1$ achieves a 40.66\% success rate. These simple results underline our cost function's innate ability for generalization across different scenes.

Figure \ref{fig:one} presents examples of planning problems where our cost outperforms that of CHOMP calibrated on the example from Figure \ref{fig:one-a}.
\captionsetup[figure]{format=withoutrule}
\captionsetup[figure]{format=withrule}

\noindent \textbf{Discussion:} In general, the need to calibrate the CHOMP cost is a direct consequence of the \textit{global optima property} (\ref{glob-optim}), which our cost satisfies by definition. Although the calibrated CHOMP cost performs reasonably well on these simple examples, the calibration process relies on our ability to cherry-pick difficult examples from the dataset to calibrate on, as the \textit{global optima property} (\ref{glob-optim}) needs to be satisfied across all examples in a dataset. However, without the inclusion of an object-specific size parameter in the loss function, CHOMP path lengths are necessarily compromised on "easier" examples when calibrating for the "hardest", to guarantee no collisions across the entire dataset. For these reasons, the CHOMP cost needs to be calibrated per scene \cite{zucker2013chomp}, while as we further demonstrate in experiments (\ref{sec:cont-plan}, \ref{sec:partial-obs}), our formulation generalizes across diverse scenes and planning setups with no need for calibration.

\begin{table*}[!htb]
\centering
\caption{Quantitative comparisons with MPNet on the Complex 3D w/o lazy state contraction}
\begin{tabular}{cccc} \toprule
 Method                      & \textbf{Success rate} & \textbf{Path length / RRT*} & \textbf{Inference speed} \\ \midrule
MPNet (0 replan)              & 34.7\%                & 1.996                       & 6.3ms                  \\ 
MPNet (1 replan)              & 42.8\%                & 2.21                        & 14ms                   \\ 
MPNet (2 replan)              & 45.7\%                & 2.354                       & 31.8ms                  \\ \midrule
\textbf{Ours (0 corrections)} & \textbf{76.2\%}       & \textbf{1.947}              & \textbf{1.35ms}        \\ \bottomrule
\end{tabular}
\label{tab:table-1}
\end{table*}

\subsection{3D planning from full-state}
\label{sec:cont-plan}

In this experiment, we test the performance of our approach against learning-based planners using a complete description (``full-state'') of the scene as input. Specifically, we compare with MPNet on the Complex 3D dataset \cite{qureshi2019motion}.

\noindent \textbf{Setup:} We train $f_{\theta}$ on $\Pi \in \mathbb{R}^{10 \times 6}$ vectorized scene descriptors, as each box has its own translation and dimensions. We randomly sample $\Pi$ in a scene of size 20, with each dimension of each of the boxes being either 5 or 10, just as in the Complex 3D training set. Further, to train $f_{\theta}$ we randomly sample start and target configurations $s, t \in \mathbb{R}^{3}$ so that there is a 50/50 breakdown between examples which would or would not collide by simply following a straight-line path. We use a simple fully connected network architecture of depth 15, with 10 highway layers\cite{srivastava2015highway} of width 256, 2 input layers of width 128, and 3 output layers of width 128. The parameters of $\cost$ and $\Theta_{p}$ were set to $s = 0.05$, $p = 2$, $\delta = 5$, and $n = 10$. 

%Further, In an attempt to measure the effect of changing $n$ on inference speed, we also trained $f_{\theta}$ on an increased scene size of 60 on \textit{Complex3D}. The effect of this change is that there now exist more \textit{trivial} solutions to the planning problem by going around multiple objects at a time. For this reason, we set $n = 3$ while keeping the other parameters the same as above.  

\noindent \textbf{Results:} Table \ref{tab:table-1} presents the performance of each method on 2000 planning problems from the unseen Complex 3D \cite{qureshi2019motion} test set. In our experiment, we measure the rate of collision-free paths, the length of the predicted paths with respect to RRT*, and the planner's inference speed. In our primary experiment seen in Table \ref{tab:table-1}, our method outperforms the learning-based component of MPNet for an arbitrary number of MPNet's replanning attempts in terms of all measured metrics.

In terms of inference speed comparison with respect to classical planners on the Complex 3D dataset, Informed-RRT* takes an average of 15.54 seconds to plan, and BIT* an average 8.86 seconds. While both BIT* and Informed-RRT* are probabilistically complete planning methods, their inference speed is much slower than our method's. Hence, we can conclude that our method is preferable for applications where rapid path planning is necessary.

%Further, to understand how $n$ relates to inference speed, in case of the scene size 60 experiment, our method achieves a success rate of 86.1\%, a length ratio of 1.466, and 0.95ms per planning problem on average. In general, we observe that the effect of $n$ on inference speed is negligible, with around $5 \times 10^{-2} \text{ms}$ per additional control point. 

%\begin{table}[!htb]
%\centering
%\caption{Inference speed comparisons on the Complex 3D dataset}
%\begin{tabular}{cccc} \toprule
% Method             &\textbf{Inference speed} \\ \midrule
%MPNet (best)         & 6.3ms                  \\ 
%Informed-RRT*        & 15.54s                   \\
%BIT*                 & 8.86s                   \\ \midrule
%\textbf{Ours (best)}  & \textbf{1.35ms}         \\ \bottomrule
%\end{tabular}
%\label{tab:table-3}

%\end{table}

\subsection{3D planning from images}
\label{sec:partial-obs}

\begin{figure}[h!]
  \centering
  \subfloat{\includegraphics[width=.23\textwidth, height=.20\textwidth]{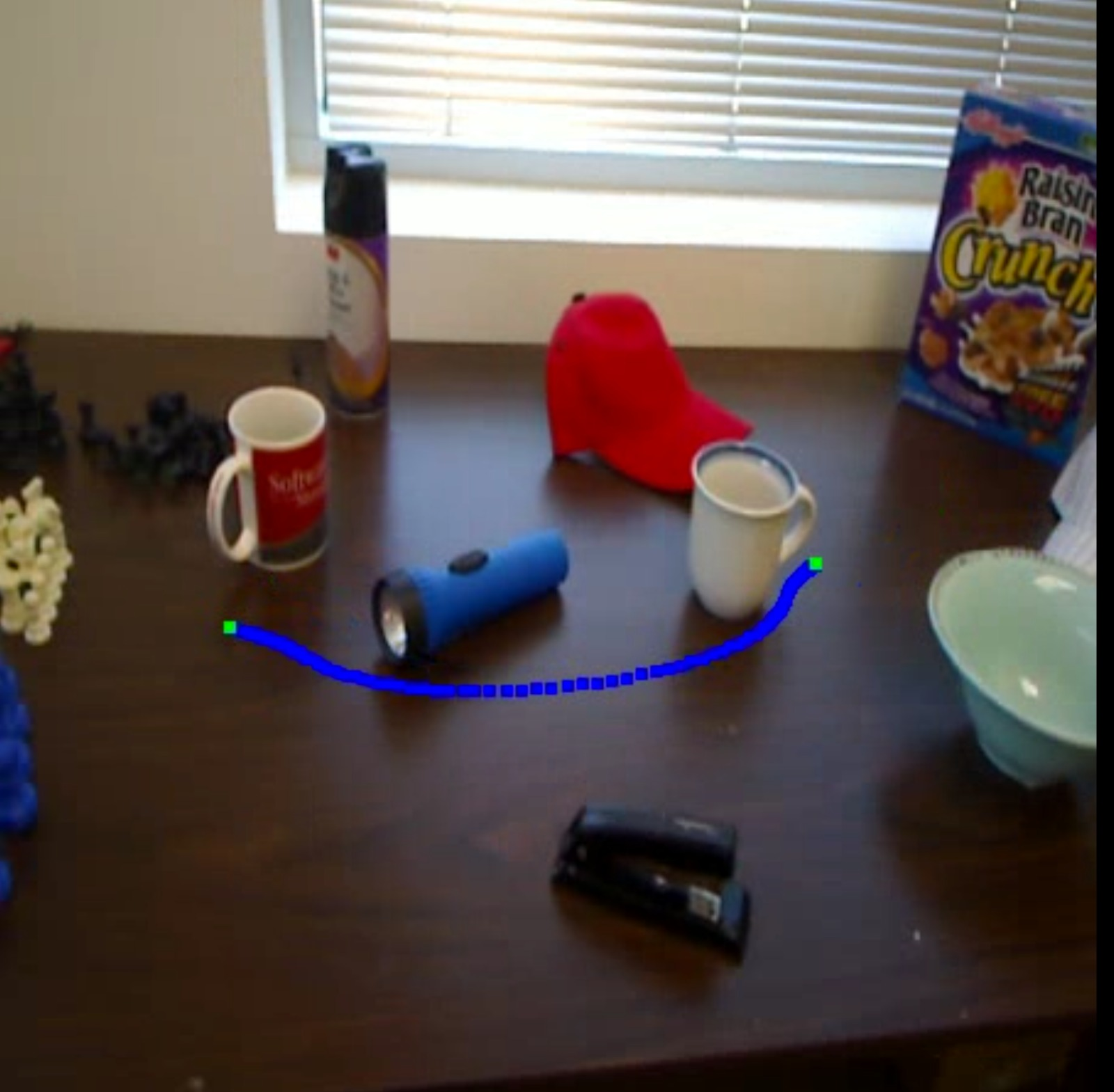}}\hfill
  \subfloat{\includegraphics[width=.23\textwidth, height=.20\textwidth]{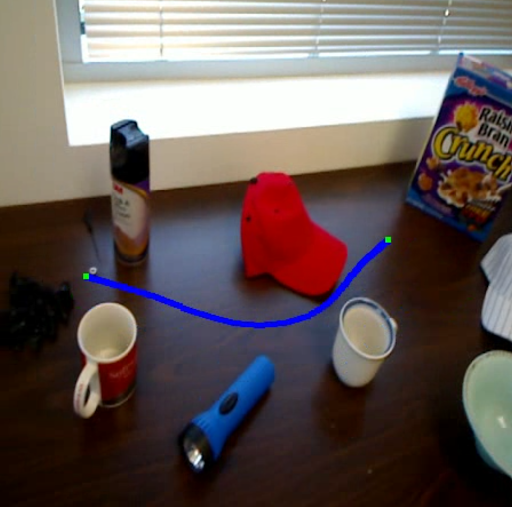}}\par
  \caption{\textbf{Predicting paths from real-world RGB-D images.} This figure shows paths on real table-top scenes from the RGB-D Scenes Dataset v.2 \cite{lai2011large}. The model is trained on purely synthetic scenes from \textit{Table-top shapes} dataset.
}
  \label{fig:real-vis}
\end{figure}

\begin{figure}[h!]
  \centering
  \subfloat{\includegraphics[width=.23\textwidth, height=.20\textwidth]{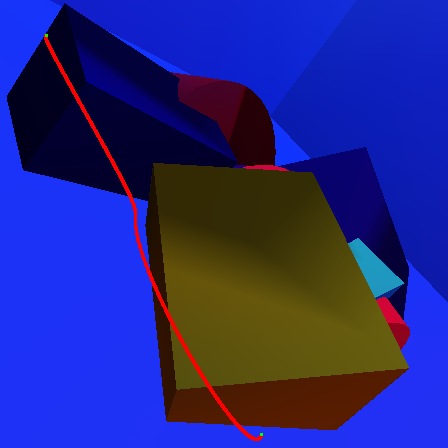}}\hfill
  \subfloat{\includegraphics[width=.23\textwidth, height=.20\textwidth]{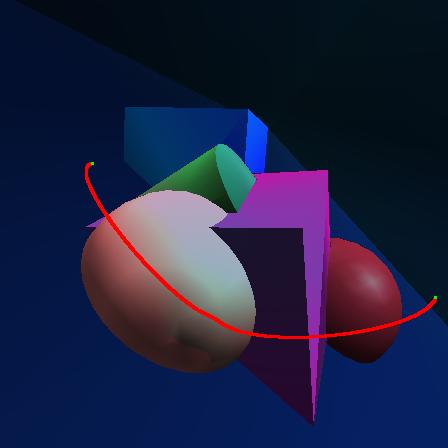}}\par
  \caption{\textbf{Predicting paths from synthetic RGB-D images.} This figure shows predicted paths on our \textit{Table-top shapes} test set.
}
  \label{fig:syn-vis}
\end{figure}

\begin{table*}[h!]
\centering
\caption{RRT* motion length (len) and success rate (succ) on 6 DoF planning problems from \textit{all-6DoF} and \textit{difficult-6DoF} test sets with $1$ms, $10$ms, $100$ms, $1$s, $10$s of planning time. Note that we set N/A where RRT* fails to find a solution to any planning problem.}
\begin{tabular}{c|cc|cc|cc|cc|cc}
\toprule
& \multicolumn{2}{c|}{\textbf{\textless1ms}} & \multicolumn{2}{c|}{\textbf{\textless10ms}} & \multicolumn{2}{c|}{\textbf{\textless100ms}} & \multicolumn{2}{c|}{\textbf{\textless1s}} & \multicolumn{2}{c}{\textbf{\textless10s}}\\
& \textbf{len} & \textbf{succ} & \textbf{len} & \textbf{succ} & \textbf{len} & \textbf{succ} & \textbf{len} & \textbf{succ} & \textbf{len} & \textbf{succ}\\
\midrule
\textit{all-6DoF}       & N/A  &  0\%  &  7.52        & 0.15\%                & 16.98             &  1.45\%           & 28.71       & 28.2\%                & 29.80 & 95.1\%\\ 
\textit{difficult-6DoF}   & N/A      & 0\%                    & N/A        & 0\%                   & 23.31         & 0.07\%                & 31.04      & 10.2\%                 & 32.4 & 91.8\%\\ \bottomrule 
\end{tabular}
\label{tab:table-dof}
\end{table*}

In this experiment, we show how our approach can be used to plan from images. Using our cost function, we train our network to predict collision-free paths conditioned on RGBD images of scenes. 

\noindent \textbf{Setup:} In this case, we have $\Pi \in \mathbb{R}_{\geq 0}^{448 \times 448}$, representing a depth image from the robot's viewpoint, with the control points of $\theta_{p}$ being in the camera frame of the scene. We use the \textit{Table-top shapes} dataset for this experiment.

The architecture we chose for $f_{\theta}$ is a ResNet-50\cite{he2016deep} backbone, followed by a $4$ layer fully-connected network of width $256$. To train our network, similarly as in \ref{sec:cont-plan}, we sample start and end configurations so that there is a $50$/$50$ breakdown between examples where a straight-line path would or would not collide. The parameters of $\theta_{p}$ and $\cost$ are set to $n = 3$, $s = 0.05$, $p = 2$, $\delta = 0$. Further, we apply thresholded perlin noise to our RGBD images, with the aim to assess the generalization of our method to real-world images.

\noindent \textbf{Results:} On $2000$ unseen examples from our synthetic test set, our method achieves an $89.05\%$ success rate on problems where a straight-line path is expected to collide and $1.39$ times longer than start-to-goal distance on problems where a straight-line path is optimal. Figure \ref{fig:real-vis} presents examples of path planning problems solved on table-top scenes from the RGB-D Scenes Dataset v.2 \cite{lai2011large}, and Figure \ref{fig:syn-vis} showcases the predicted paths on our \textit{Table-top shapes} test set.

%TODO: ADD synthetic photos

\subsection{6 DoF planning}
\label{sec:dof}

\begin{figure}[t!]
  \centering
  \includegraphics[width=\columnwidth]{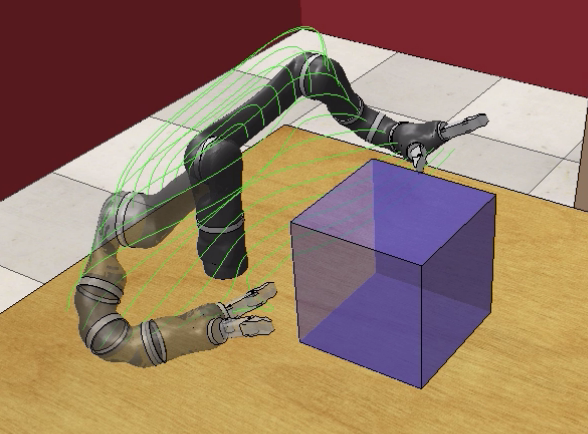}
  \caption{\textbf{Controlling a robotic manipulator.} Using Ivy \cite{lenton2021ivy}, we demonstrate a 6 DoF Kinova Mico robotic manipulator utilizing our proposed method for planning around a box. The arm motion is shown in \textbf{\color{green} green}.}
  \label{fig:end-eff}
\end{figure}

This experiment demonstrates how our approach can be used to plan motions for a 6-DoF Kinova Mico arm.

\noindent \textbf{Setup:} We train our method by sampling random manipulator start and target configurations and regressing to spline joint angles. We set $s = 0.05$, $p = 2$, $\delta = 0$, and $n = 3$. To compute our cost, we integrate it over the full manipulator motion in Cartesian space by uniformly sampling both through time and between the link positions. We use the same network architecture as in \ref{sec:cont-plan}. We compare our method with OMPL's\cite{sucan2012the-open-motion-planning-library} RRT*\cite{karaman2011sampling} on a test set of $2000$ planning problems from \textit{all-6DoF} and on a test set of $2000$ planning problems from \textit{difficult-6DoF}. Further, we showcase the use of our method for finding good initial planning solutions for downstream optimisation. We achieve this by comparing the path initialisation obtained from our network against linear interpolation in the start and target joint angles on the \textit{difficult-6DoF} dataset. We perform several gradient steps using the collision component of the cost, and compare the resulting success rates.

\noindent \textbf{Results:} Our method achieves a $56\%$ success rate on \textit{difficult-6DoF} with a $26.72$ motion length, and a $73\%$ success rate on \textit{all-6DoF} with a $25.2$ motion length. We measure motion length by sampling anchor points on the arm and computing their distances across time.  Our method's planning time per problem is 0.95ms. We compare with RRT* by letting RRT* plan up to 1ms, 10ms, 100ms, 1s, and 10s on the same problems. The results of RRT* performance can be seen in Table \ref{tab:table-dof} and an example trajectory of our method is in Figure \ref{fig:end-eff}. Based on the results from Table \ref{tab:table-dof}, although RRT* can catch up with our method in terms of success rate within 10s of planning time, the resulting RRT* planner motions are longer than our method's. For a planning setup up to 1s, our planner can provide superior results both in terms of length and success rates. Overall, our approach consistently provides superior length per planning time and success rate per planning time ratios.

\begin{figure}[t!]
  \centering
  \includegraphics[width=0.7\columnwidth]{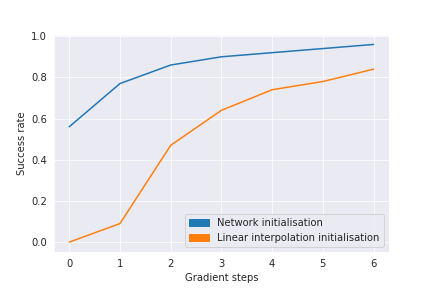}
  \caption{\textbf{A network trained using our method provides superior path initialisations} compared with linear interpolation in the joint angles.}
  \label{fig:post-grad-steps}
\end{figure}

Further, \autoref{fig:post-grad-steps} demonstrates the use of our method for good path initialisations. We observe that after $6$ gradient steps on the paths provided by our network, the planning solutions were close to optimal in terms of success rate. While our method can be used to optimise paths initialised with linear interpolation, more gradient steps are needed to achieve similar success rates.

\section{Conclusion}
% \vspace{-2mm}
In this paper, we presented an optimal cost function for learning to find the shortest collision-free paths from images. The key to our approach is a novel cost formulation which guarantees collision-free shortest paths at the optimum. Our experimental results demonstrate that our method outperforms other optimization-based planners, performs on par with supervised learning based-planners, and is effective at planning in higher-dimensions such as on a 6 DoF robotic manipulator.

{\small
\bibliographystyle{ieee_fullname}
\bibliography{ms}
}



\section*{Supplementary material}

# 1 Appendix A: Collision penalty satisfies required properties

We show that $c(O, \theta_p) = \sum_{o \in O} \mathbb{1}_o(\theta_p) * R(o)$ satisfies *minimum property*, *non-colliding property*, and the *global optima property*.

*Proof.*
Take an arbitrary path planning problem with obstacle set $O$, $s \in \mathbb{R}^d$, and $t \in \mathbb{R}^d$, with an optimal non-colliding shortest path parameterized by $\theta_{opt} \in \Theta_p$. Further, let $\forall \theta_p \in \Theta_p$: $S(\theta_p, O)$ be defined to provide a set of path segments that are colliding with objects in the scene. This means that for any $\theta_p \in \Theta_p$, we will have:

$$\theta_p \text{ does not collide} \Leftrightarrow S(\theta_p, O) = \emptyset \Leftrightarrow \forall o \in O : \mathbb{1}_o(\theta_p) = 0 \tag{1}$$

To show *non-colliding property*, by the definition of $c$ and (1), for an arbitrary non-colliding path $\theta_p \in \Theta_p$, we will necessarily have $c(O, \theta_p) = 0$. Now, to show that for any $\theta_p \in \Theta_p$: $c(O, \theta_p) = 0 \Rightarrow \theta_p$ does not collide, we know that $\forall o \in O : R(o) > 0$. Therefore, if we assume $c(O, \theta_p) = 0$, we must have $\forall o \in O : \mathbb{1}_o(\theta_p) = 0$ and so $\theta_p$ does not collide by (1). We have shown that $c$ has the *non-colliding property*.

To show *minimum property*, we note that both $R$ and $\mathbb{1}$ are non-negative, and hence $c$ is non-negative. By *non-colliding property*, because $\theta_{opt}$ is non-colliding, we have $c(O, \theta_{opt}) = 0$. We have shown *minimum property*.

For *global optima property*, take an arbitrary $\theta_p \in \Theta_p$ and consider two cases:

**Case 1:** $l(\theta_{opt}) < l(\theta_p)$
In this case, $l(\theta_{opt}) - l(\theta_p) < 0$, so *global optima property* holds by the definition of $c$. Particularly, because $c$ is always non-negative, as we saw in the proof of the *minimum property*.

**Case 2:** $l(\theta_{opt}) \geq l(\theta_p)$
First, suppose that the path given by $\theta_p$ does not collide with any objects in $O$. By the assumption of this case, we have $l(\theta_{opt}) \geq l(\theta_p)$ and so necessarily also $l(\theta_{opt}) = l(\theta_p)$. Otherwise, $\theta_p$ would have yielded a shorter path than $\theta_{opt}$, which is a contradiction with $\theta_{opt}$ being the optimal path. Hence, by *non-colliding property* and $\theta_p$ being non-colliding, we have:

$$c(O, \theta_p) = l(\theta_{opt}) - l(\theta_p) = 0 \tag{2}$$

which implies that *global optima property* holds in this case.

Now in turn, suppose that the path $\theta_p$ does collide with objects in $O$. Since we defined a loss function which is not supervised by information about $\theta_{opt}$, we do not have access to $\theta_{opt}$ and hence to $l(\theta_{opt})$. Because of this limitation, we

will now aim to derive an upper bound for $l(\theta_{opt}) - l(\theta_p)$ by defining an oracle transformation $T$, such that:

$$\forall \theta_p \in \Theta_p : S(T(\theta_p, O), O) = \emptyset \qquad (3)$$

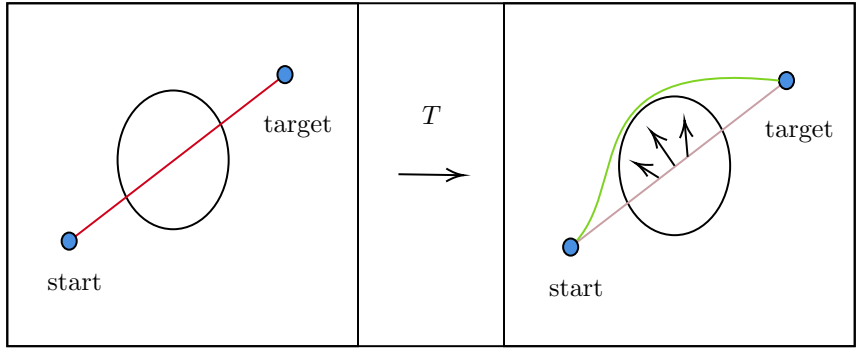

Figure 1: The $T$ transformation transforms a colliding path segment into a non-colliding segment.

Figure 1 demonstrates an example a valid $T$ transformation. Note that we must have $l(T(\theta_p, O)) \geq l(\theta_{opt})$ as $\theta_{opt}$ is shortest possible. In practice, the exact definition of $T$ could be task space specific and we would be able to prove *global optima property* and consequently optimize $f_\theta$. However, defining $T$ to project the colliding segments of the path $\theta_p$ marginally beyond its respective colliding objects, or possibly on the respective object bounding spheres is sufficient.

Assuming the definition of $T$ as above, we may write:

$$T(\theta_p, O) = (\theta_p \setminus (\bigcup_{\theta'_p \in S(\theta_p, O)} \theta'_p)) \cup (\bigcup_{\theta'_p \in S(\theta_p, O)} T(\theta'_p, O)) \qquad (4)$$

Since $l$ is a measure on $\Theta_p$, by countable additivity, we have:

$$\begin{aligned} l(T(\theta_p, O)) &= l(\theta_p) - \sum_{\theta'_p \in S(\theta_p, O)} l(\theta'_p) + \sum_{\theta'_p \in S(\theta_p, O)} l(T(\theta'_p, O)) \\ &= l(\theta_p) + \sum_{\theta'_p \in S(\theta_p, O)} l(T(\theta'_p, O)) - l(\theta'_p) \end{aligned} \qquad (5)$$

Now, we may consider $l(T(\theta'_p, O)) - l(\theta'_p)$ for arbitrary $\theta'_p \in S(\theta_p, O)$. Let $o \in O$ be the object that $\theta'_p$ collides with. Then, we will have:

$$R(o) \geq l(T(\theta'_p, O)) \geq l(T(\theta'_p, O)) - l(\theta'_p) \qquad (6)$$

Hence, we have from 5, 6:

$$l(\theta_p) + \sum_{o \in O} \mathbb{1}_o(\theta_p) * R(o) \ \geq \ l(T(\theta_p, O)) \geq l(\theta_{opt}) \tag{7}$$

From 7 and definition of $c$, then directly follows *global optima property*. □

## 2  Appendix B: Training specifics

The general network architecture we use in our experiments is depicted in Figure 2. Note that we tailor both the input network, the backbone, and the fully-connected network to the planning dataset at hand. For example, if we plan based on images, we might use a CNN in place of the backbone network; otherwise, a simple fully-connected network might suffice.

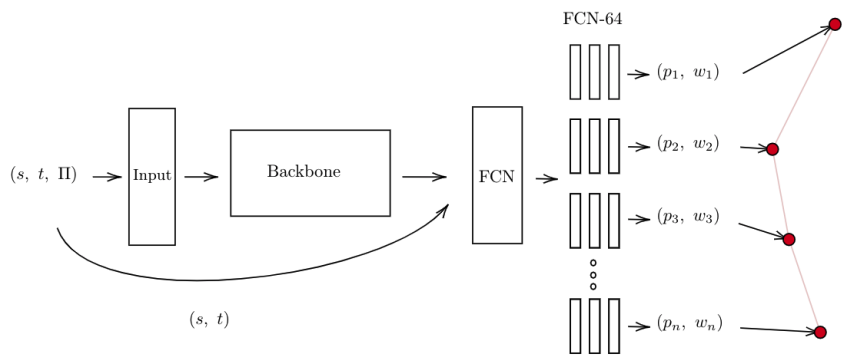

Figure 2: General network architecture used in our experiments.

In the paper, we trained two different network architectures. One for the *3D planning from images* experiment and one for the *3D planning from full-state* and *6 DoF planning* experiments. The tailored components of the general network are specified in Table 1.

| Network  | 3D planning from images   | 3D planning from full-state, 6 DoF    |
|----------|---------------------------|---------------------------------------|
| Input    | Conv3, 2x                 | MLP - 2 layer, width 128              |
| Backbone | ResNet-50                 | Highway layers - 10 layers, width 256 |
| FCN      | MLP - 4 layer, width 256  | MLP - 3 layer, width 256              |

Table 1: Tailored components of the general network for the planning tasks in our paper.

# 3 Appendix C: Control point impact on inference speed

In an attempt to measure the effect of changing $n$ on inference speed, we also trained $f_\theta$ on an increased scene size of 60 on *Complex3D*. The experiment was carried out the same way as the experiment in the *3D planning from full-state* section. The effect of this change is that there now exist more *trivial* solutions to the planning problem by only going around multiple objects at a time. For this reason, we set $n = 3$ while keeping the other parameters the same as for the scene size 20 comparison from section *3D planning from full-state*.

Table 2: Results of our method on the Complex 3D w/ scene size 60

| Method | Success rate | Length / RRT* | Inference speed |
|---|---|---|---|
| **Ours (0 corrections)** | 86.1% | **1.466** | **0.95ms** |
| **Ours (1 correction)** | **91.45%** | 1.487 | 1.8ms |

We observe that the effect of $n$ on inference speed is negligible, around $5 \times 10^{-2}$ms per $n$.

# 4 Appendix D: Continuous planning results visualized

Figure 3 visualizes the results of our experiment from the *3D planning from full-state* section. In particular, we visualize the experiment where MPNet was planning without *lsc*.

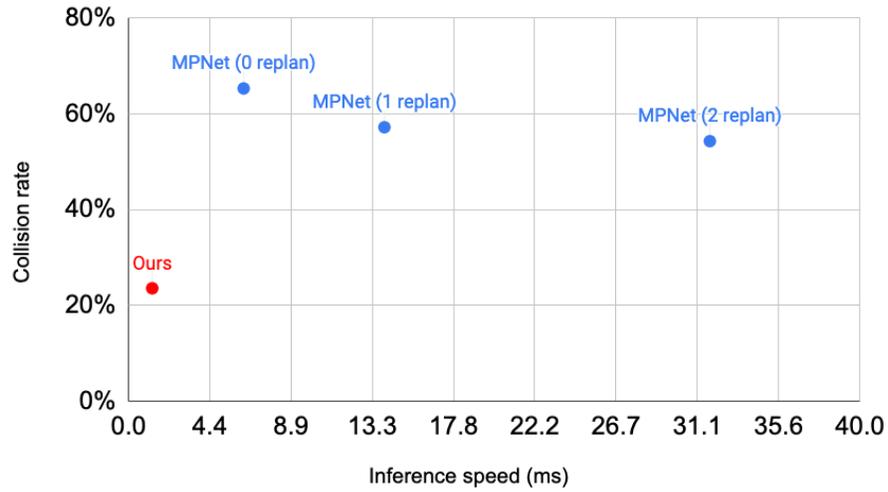

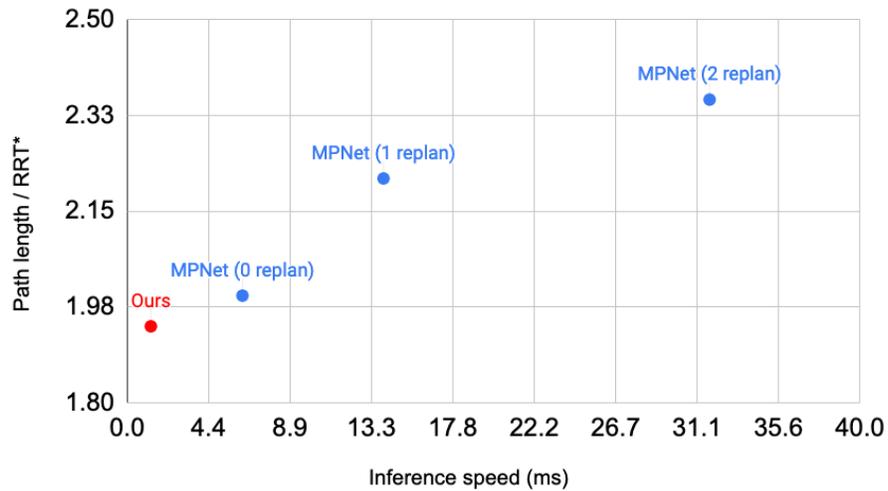

Figure 3: The results from Table 1 in section *3D planning from full-state* indicate that our method outperforms MPNet both in terms of planned path success rate and length.

## 5   Appendix E: 6 DoF planning cost

This appendix shows how our cost is computed in the *6 DoF planning* experiment. Figure 4 showcases how the manipulator is sampled to calculate our cost function. To compute the cost, we compute it for each anchor's trajectory (red dotted lines in Figure 4) and then minimize the sum over all trajectories.

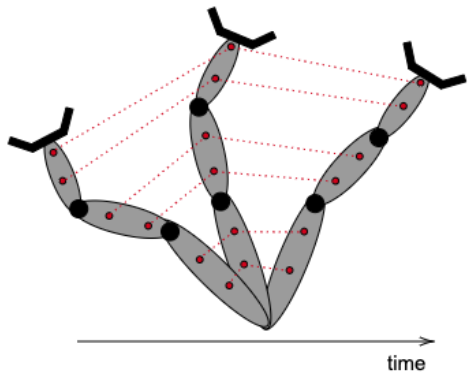

Figure 4: Illustration of how the manipulator is sampled when computing the cost function for the *6 DoF planning* experiment.

## 6   Appendix F: Synthetic examples

Figure 5 presents several examples of path planning problems solved by our method in the *Table-top shapes* test set.

## 7   Appendix G: Vectorized scene descriptor

In experiments *3D planning from full-state* and *6 DoF planning*, we use $\Pi \in \mathbb{R}^{k \times d}$ vectorized scene descriptors as the input to our method. Here, $k$ refers to the number of objects in the scene, and $d$ is the dimension of their properties. For example, if we are planning solely around axis-aligned boxes, we can set $d = 6$, with each box being represented by its translation and dimensions. If we add rotation, we may set $d = 11$ and include each of these boxes' quaternion.

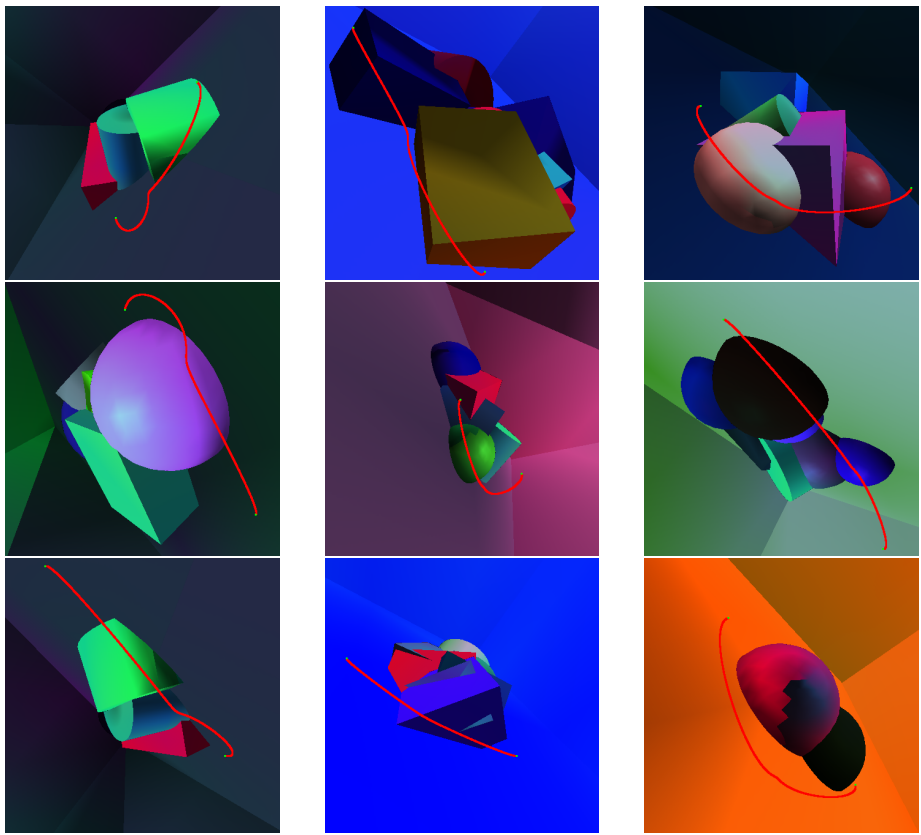

Figure 5: Examples of $f_\theta$ paths solved in our *Table-top shapes* synthetic test set.



\end{document}